\pdfoutput=1

\documentclass[11pt]{article}

\usepackage{emnlp2021}

\usepackage{times}
\usepackage{latexsym}

\usepackage[T1]{fontenc}

\usepackage[utf8]{inputenc}

\usepackage{microtype}

\usepackage{amssymb}
\usepackage{booktabs}
\usepackage{comment}
\usepackage[acronym]{glossaries}
\usepackage{graphicx}
\usepackage{todonotes}

\glsdisablehyper
\newacronym{NMI}{NMI}{normalized mutual information}

\title{Discrete representations in neural models of spoken language}

\author{ Bertrand Higy\\
  Cognitive Science and AI\\
  Tilburg University \\
  \small{\texttt{b.j.r.higy@tilburguniversity.edu}}\\\And
  Lieke Gelderloos\\
  Cognitive Science and AI\\
  Tilburg University \\
  \small{\texttt{l.j.gelderloos@tilburguniversity.edu}}\\\AND
  Afra Alishahi\\
  Cognitive Science and AI\\
  Tilburg University \\
  \small{\texttt{a.alishahi@tilburguniversity.edu}}\\\And
  Grzegorz Chrupała \\
  Cognitive Science and AI\\
  Tilburg University \\
  \small{\texttt{grzegorz@chrupala.me}}}

\begin{document}
\maketitle

\begin{abstract}
  The distributed and continuous representations used by neural
  networks are at odds with representations employed in linguistics,
  which are typically symbolic. Vector quantization has been proposed
  as a way to induce discrete neural representations that are closer
  in nature to their linguistic counterparts. However, it is not clear
  which metrics are the best-suited to analyze such discrete
  representations. We compare the merits of four commonly used metrics
  in the context of weakly supervised models of spoken language.  We
  compare the results they show when applied to two different models,
  while systematically studying the effect of the placement and size
  of the discretization layer.  We find that different evaluation
  regimes can give inconsistent results. While we can attribute them
  to the properties of the different metrics in most cases, one
  point of concern remains: the use of minimal pairs of phoneme triples as
  stimuli disadvantages larger
  discrete unit inventories, unlike metrics applied to complete
  utterances. Furthermore, while in general vector quantization
  induces representations that correlate with units posited in
  linguistics, the strength of this correlation is only moderate.
\end{abstract}

\section{Introduction}
\label{sec:intro}

The dominant machine learning paradigm for processing spoken language
is based on neural network architectures, such as recurrent nets and
transformers, inducing a hierarchy of hidden representations which are
distributed and continuous. In contrast, human history has repeatedly
seen the discovery and wide adoption of discrete, symbolic
representations of speech in the form of writing. These systems commonly
represent basic units of language such as morphemes, syllables or phonemes
while discarding other information contained in the speech signal such as
emotion or speaker identity.

Symbolic representation of speech has proven tremendously useful for storage
and transmission of information, and it also plays a crucial role in
systems dealing with spoken language such as spoken dialog systems: these
typically employ an automatic speech recognition (ASR) module to transcribe the
speech signal into written form, which is then used as input to upstream
language understanding modules. While some attempts have been made to train
such systems end-to-end, pipelines are still very competitive
\citep[see][]{haghani_audio_2018, higy2020textual}, which strengthens the point
that a symbolic encoding of spoken language contains most of the information relevant for this task.

It may thus be desirable to incorporate similar representations in neural
architectures. Accordingly, multiple efforts have been made to design neural
networks with discrete hidden representations and to apply them to spoken
language data. This is evident in the recent editions of the ZeroSpeech
challenge \citep{dunbar2019zero} on unit discovery, which have featured many
such approaches. Specifically, Vector Quantization (VQ) has proven to be a
simple and effective method to induce discrete neural representations
\citep[e.g.,][]{vanoord2017neural,harwath2020learning,chung2020vector, liu_cross-modal_2021}.  
VQ layers are added to neural architectures in order to map continuous
activation vectors onto a finite set of discrete units, often referred
to as {\it codes}, via a dictionary (or {\it codebook}) associating these codes with
their vector embeddings; the number of entries in the codebook is the {\it codebook size}. 
Such symbolic codes have been claimed to 
correspond to phonemes and/or words. What is still lacking though is a 
detailed analysis of how much the reported equivalence is affected by details of
the architectures such as the size and placement of the VQ layers, 
learning objectives and dataset, as well as by the evaluation metrics used to
quantify it. The present study aims to fill this gap.

We study two approaches to modeling spoken language: learning driven
by language-internal structure, and learning driven by grounding in
the extra-linguistic world. These two approaches are exemplified by
two types of models with VQ layers: the self-supervised model for unit
discovery of \citet{van2020vector}, and a visually grounded model
similar to \citet{harwath2020learning}. The datasets used to train
each model, Zerospeech 2020 \citep{dunbar_zero_2020} and Flickr8K 
\citep{harwath2015deep,rashtchian-etal-2010-collecting} are also typical of the task
they are used for. Using these two models as our test cases, we 
systematically investigate the impact of the following factors: (i)
the codebook size for the VQ layer, and (ii) the level of placement of
the VQ layer. Furthermore, we apply and check the consistency across
four different metrics for evaluating the correspondence of the
representations with phonemes.

\paragraph{Findings}
The self-supervised model shows high variability, but with some of the
evaluation metrics (especially ABX and RSA, see
Section~\ref{sec:evaluation-methods} for the definition of the metrics) there is a trend for better
correspondence to phonemes for smaller codebook sizes (32-128).  The
visually grounded model, on the other hand, generally shows the
closest match with phonemes for the {\it largest} codebook sizes (512
or 1024) when evaluated on full utterances. In contrast, smaller
codebooks score higher for the short utterance segments used by the
ABX metric (i.e. minimal pairs of phoneme triples). We also observe
inconsistencies in the relative performance of VQ layers placed at different
levels in the model for RSA vs. the other metrics. As discussed in
Section~\ref{sec:discussion}, we attribute those inconsistencies to the
properties of the different metrics. Thus, the
conclusions drawn using a single metric, even a widely used one like
the ZeroSpeech ABX metric, should be treated with caution, and
further corroborated.

\section{Related work}
\label{sec:related}

The domain of speech processing has seen recent efforts to modify
existing neural architectures to enable the induction of discrete latent
representations. These developments are promising for boosting
performance, improving interpretability, and modeling the acquisition and
processing of linguistic knowledge in humans.

\begin{table*}[htb]
    \caption{A comparison of studies of VQ-based speech models. RAW (Reconstructing Audio Waveform); VAE (Variational Autoencoder); APC (Autoregressive Predictive Coding); CPC (Contrastive Predictive Coding); DC (Diagnostic Classifier).}
\vspace{-10pt}
\label{tab:studies}
\begin{center}
{\footnotesize

\begin{tabular}{p{3.2cm}p{2.1cm}p{1.5cm}p{4.0cm}p{3.0cm}}
\toprule
{\bf Study}&{\bf Objective}&{\bf Model} & {\bf Analysis} & {\bf Manipulated factors} \\
\midrule
\citet{vanoord2017neural}  & RAW & VAE & Phoneme (majority label) & None \\
\citet{chorowski_unsupervised_2019} & RAW & (V)AE, VQ-VAE & Phoneme (ABX, frame-wise DC), gender, speaker &
VAE latent dimensions \\
\citet{chung2020vector} & Predicting next frame & APC & Phoneme (frame-wise DC), speaker &
VQ layer number, position, codebook size \\
\citet{van2020vector} & RAW, predicting future frames & VQ-VAE, VQ-CPC &
Phoneme (ABX), speaker, voice conversion  & None
\\
\citet{harwath2020learning} & Visual grounding & ResDAVEnet
                            & Phoneme (ABX), bitrate, word (F1 scores) & VQ layer number, position, training regime \\
\bottomrule
\end{tabular}
}
\end{center}
\vspace{-5pt}
\end{table*}

Applying Vector Quantization (VQ) techniques for this purpose was
pioneered by \citet{vanoord2017neural}, who propose generative models
based on the variational auto-encoder (VAE) architecture and use VQ to
induce discrete latent representations. They apply this method to
images, videos, and speech, and show that the models can learn discrete latent representations
without supervision. When applied to raw
speech, the VQ-VAE architecture learns high-level discrete
representations that are invariant to low-level features of the audio
signal such as prosody and speaker identity, and mostly encode the
content of the speech. Classification of the discrete representations
into phoneme classes (based on majority ground truth label) suggests
they capture phonemes to some extent.

Learning discrete instead of (or in addition to) continuous
representations can facilitate unit discovery in unsupervised models
of speech. In the 2015 Zero Resource Speech Challenge
\citep{versteegh2015zero}, \citet{badino2015discovering} present a
binarized auto-encoder, certain variants of which outperform its
continuous counterpart.  In the 2017, 2019, and 2020 editions of this
challenge \citep{dunbar_zero_2017, dunbar2019zero, dunbar_zero_2020},
following the work of \citet{vanoord2017neural}, several models
include at least one VQ layer \citep[see
e.g.][]{chorowski_unsupervised_2019, eloff_unsupervised_2019,
  tjandra_vqvae_2019, tjandra_transformer_2020}. These studies
demonstrate the benefit of using VQ layers for phoneme classification
and for learning speaker-invariant representations, focusing on the
ABX phoneme discrimination metric to evaluate the encoding of phonemic
information. However, little analysis on the impact of the size and
configuration of the employed VQ layers is
performed. For example, \citet{van2020vector}
train a VQ-VAE model for reconstructing audio waveforms,
and a VQ-CPC (Contrastive Predictive Coding) model for predicting
future acoustic units. They evaluate these architectures on the ABX
phoneme discrimination task, voice conversion and speaker classification and show that VQ-CPC
performs better than VQ-VAE overall, but they do not manipulate the
configuration or the dimension of the VQ layers.

\citet{chung2020vector}, however, report the impact on phoneme and
speaker classification of systematic manipulation of VQ-related
factors. They train an Autoregressive Predictive Coding (APC) model to
predict upcoming frames, and use VQ as a methodology to limit the
model's capacity. Using a frame-wise diagnostic classifier (namely
linear logistic regression), they show that under restricted
configurations (only one VQ layer inserted at the top), phoneme
prediction using discrete representations improves over using
continuous representations learned by an APC model without VQ. In this
configuration, larger codebook sizes lead to better performance in phoneme
classification but not in speaker identification.

In contrast to the work cited above, which discusses uni-modal speech
models, \citet{harwath2020learning} use VQ layers within the
setting of learning spoken language via grounding in the visual
modality, where the speech signal is associated to images \citep[for
an overview of visually grounded speech models,
see][]{DBLP:journals/corr/abs-2104-13225}.  They hypothesize that
discrete representations learned by such models are more likely to
capture higher-level semantic information. Their analyses suggest that
the trained multimodal models can learn discrete linguistic units at
both word and sub-word levels, with quantization layers inserted at
the lower levels of the network showing correspondence to sub-word
(i.e.\ phonemic) units, and those inserted at the higher level
corresponding to word-level units. Analyses are based on the
Zerospeech ABX metric for phoneme encoding and F1 scores for word
detection.
Similarly, \citet{liu_cross-modal_2021} propose a framework based on VQ to discover discrete concepts in models of visually grounded language trained on video and text, video and audio or image and audio. Evidence of a correspondence between the learned concepts and visual entities/actions or words are given but no detailed analysis is performed.

When evaluating the encoding of phonemic information in continuous
representations from models of spoken language, recent work has shown
that different metrics may yield different outcomes.
\citet{chrupala-etal-2020-analyzing} show that representational
similarity analysis (RSA) and diagnostic classifier (DC) applied to
pooled representations disagree with the results of DC applied on
local representations for RNN-based architectures, while they are all
in agreement when applied to transformer-based representations.
\citet{algayres:hal-02977539} compare the aforementioned ABX metric
and the mean-average-precision (MAP) metric (which uses
representations to predict whether two stimuli have the same
ground-truth label) to each other and to a downstream frequency
estimation task. Performance on the three metrics is correlated, but
not to a high degree, and marked discrepancies are found for
particular models.

Table~\ref{tab:studies} summarizes some of the representative studies
that use VQ layers and their specifications and reported analyses. As
can be seen from this summary, existing work on learning VQ-based
discrete representations does not easily lead to a coherent picture
due to the wide range of the training objectives and modeling architecture
they use, the analyses they perform, the evaluation metrics they
employ and the VQ-related factors they manipulate.  In this paper, we
aim to provide this overview by employing different discretized
speech modeling approaches and consistently comparing architectural 
parameters and evaluation metrics.

\section{Methods}

\subsection{Vector quantization}
We investigate evaluation metrics for the analysis of discrete
representations induced through vector quatization.  Since phoneme
classification/identification has been the dominant analysis task for
discrete representations of speech, we use this as our main task.  We
do so through the specific case of speech representations learned by
two different models: a self-supervised model of speech
trained to reconstruct the audio waveform, and a visually-supervised
model of spoken language which maps audio representations of spoken
utterances and visual representations of their corresponding images to
a shared semantic space.  Both models employ VQ layers in their
architecture to induce discretized representations.  A VQ layer takes
as input a continuous distributed representation in the form of a
vector $h \in \mathbb{R}^d$, and returns the closest of $K$ prototype
vectors contained in a trainable codebook $\{e_1, e_2, \ldots, e_K\}$
where $e_i \in \mathbb{R}^d$. For a sequence of continuous vectors
$(h_1, h_2, \dots, h_n)$ the discrete codes are given by the sequence
of indices of the prototype vectors returned by the VQ layer.  Since
the $\arg\max$ operation needed to select the nearest vector is not
differentiable, the gradient for backpropagation is approximated by
using the straight-through estimator \citep{bengio2013estimating},
which replaces each non-differentiable operation with the identity
function for the backward pass. For further details, consult
\citet{vanoord2017neural}.

\subsection{Target models}
\paragraph{Self-supervised}
Our self-supervised model is the VQ-VAE model introduced in
\citet{van2020vector}.\footnote{We use the authors' implementation
  available at \href{https://github.com/bshall/ZeroSpeech}{github.com/bshall/ZeroSpeech}.}
The model consists of an {\it encoder} built out of a stack of five
convolutional layers, a {\it bottleneck} comprising a linear
projection and a VQ layer, and a {\it decoder} comprising an embedding
layer and a stack of upsampling and recurrent layers; the decoder
attempts to reconstruct the original waveform. For details of the
architecture, see \citet{van2020vector}. In the experiments reported
here, we vary the size of the codebook, but keep the placement of the
VQ layer constant as the encoder contains only one fully connected
layer after which the VQ layer can be placed.

\paragraph{Visually-supervised}
A visually-supervised model of spoken language with discrete
representations was introduced by \citet{harwath2020learning}: they
adapted an existing model \citep{harwath2020jointly} by inserting one
or more VQ layers within the speech encoder stack. We similarly adapt
the architecture used in \citet{Merkx2019} and
\citet{chrupala-etal-2020-analyzing} by inserting a single VQ layer at
one of three levels: either following the first, second or third GRU
layer of the speech encoder. In addition to VQ layer placement, we
also vary the size of the codebook. Note that unlike
\citet{harwath2020learning} we do not use a pre-training stage which
by-passes the VQ layers; rather, we train the complete network from
scratch. This model thus consists of an image encoder, which takes as
input image features extracted via a pre-trained ResNet-152 model
\citep{he2016deep} and maps these features via a learned affine
transform into a joint visual-language space.  The audio input are
MFCC features with total energy and delta and double-delta
coefficients with combined size 39.  The speech encoder consists of
one 1D convolutional layer (with 64 output channels) which subsamples
the input by a factor of two, four bidirectional GRU layers (each of
size 2048), with a VQ layer inserted between a single pair of GRU
layers. This stack is followed by a self-attention-based pooling
layer. The objective function is a version of the triplet loss with
negative examples from the current batch. The model is trained with
the Adam optimizer \citep{kingma2014adam} with a cyclical learning
rate schedule \citep{smith2017cyclical}.

\subsection{Evaluation methods}
\label{sec:evaluation-methods}

Different evaluation methods have been used for analyzing the nature of
information captured by VQ-based discrete representations in different studies.
We present here a thorough examination of their formal similarities and
differences as well as their sensitivity to different conditions. In this section we
introduce the methods commonly used to evaluate the learned representations.


\citet{chrupala-etal-2020-analyzing} study the effect
representation scope, i.e.\ activation vectors retrieved at the level
of frames (local) or pooled over whole utterances (global), concluding
that it can affect results. Following their recommendations, and for
the sake of simplicity, we include one measure for each scope. Since
their findings suggest that local RSA lacks sensitivity, we use local
DC (which is also the most widely used) as well as global RSA in our
experiments.  Global RSA has the double advantage of not using any
trainable parameters and not requiring any alignments.

\paragraph{Normalized Mutual Information (NMI)}

An information-theoretically motivated measure of the association
between two random variables is
mutual information. In the
general case of vector-valued neural representations, computing mutual
information with the target annotation is intractable. In the special
case where the representation is discrete-valued, we can use the
standard empirical estimate. Given discrete random variables $X$ with
image $\mathcal{X}$ and $Y$ with image $\mathcal{Y}$ (i.e.\ frame-wise
codes and phoneme labels in our case), the mutual information $I(X;Y)$
is
\begin{equation}
  I(X;Y) = \sum\limits_{x \in \mathcal{X}} \sum\limits_{y \in
    \mathcal{Y}}
  P(x,y) \log \frac{P(x,y)}{P(x)P(y)}
  \label{eq:mi}
\end{equation}

It is often more informative to use  mutual information normalized
by the  arithmetic mean of the entropies of the two random variables:
\begin{equation}
  \textrm{NMI}(X;Y) = 2 \frac{I(X;Y)}{H(X) + H(Y)}
  \label{eq:nmi}
  \end{equation}
where $H(X)$ is the entropy of $X$. This definition of \gls{NMI} is
equivalent to the V-measure \citep{rosenberg_v-measure_2007}.

\paragraph{Diagnostic Classifier (DC)}
A diagnostic model, also known as a {\it probe}, is a classifier or
regressor trained to predict some information of interest (such as a
linguistic annotation) given a neural representation. To the extent
that the model successfully predicts the annotation, we conclude that
the neural representation encodes this information. Informally, such
a diagnostic classifier can be seen as quantifying the amount of
easily-accessible -- or in the extreme case, linearly decodable --
information about the target annotation \citep[][among
others]{adi2016fine,alishahi-etal-2017-encoding,hupkes2018visualisation,conneau-etal-2018-cram}.

As argued by
\citet{pimentel-etal-2020-information}, without the qualification that
information be easily accessible, probing should aim to
approximate the mutual information between the neural representation
and the target annotation, and thus should use the best-performing
probe possible.  Furthermore, it is not possible for the neural
representation to contain more information about the target annotation
than the source utterance itself, due to the information processing
inequality, and thus, in the general case,
probing with an unrestricted classifier is not a well-founded
exercise. In the special case of probing a discrete-valued variable
(as is the case in our study)
the situation is simpler:
the accuracy of a linear classifier is closely related to
the empirical estimate of the mutual information between the
representation and the target annotation; see the formal argument in
Appendix~\ref{sec:loss-diag} as well as our empirical results.

\paragraph{Representational Similarity Analysis (RSA)}
RSA is a second-order technique originating in neuroscience
\citep{kriegeskorte2008representational} where similarities between
pairs of stimuli are measured in two representation spaces: e.g.\
neural activation pattern space and a space of symbolic linguistic
annotations such as sequences of phonemes or syntax trees.  The
correlation between these pairwise similarity measurements quantifies
how much the two representations are aligned. This approach requires a
similarity or distance metric for pairs of stimuli within each
representation space, but does not need a way of mapping from one
space to the other. It generally does not have any trainable
parameters. As a consequence, it is sensitive to the 
purity of the representation with regard to the variable of interest:
unlike DC, the RSA metric will penalize representations for
encoding any information unrelated to the target variable.
See for example \citet{bouchacourt-baroni-2018-agents,
  chrupala-alishahi-2019-correlating, abnar-etal-2019-blackbox,
  abdou-etal-2019-higher, chrupala-etal-2020-analyzing,
  fereidooni2020understanding, davis2020discourse} for uses in NLP and
speech processing.

When the neural representations of the stimuli evaluated are sequences
of vectors, we need to make a choice regarding how to measure
similarities or distances between them.
Here we focus on neural representations which take
the form of sequences of symbolic codes, which makes measuring
distances simple: a natural choice is the Levenshtein edit distance
normalized by the length of the longer string. We can thus apply the
same edit-distance metric on both the neural representations and on
the reference sequences of phonemes or words (for efficiency we
collapsed code repetitions).

%
%
\paragraph{ABX discriminability (ABX)}
The ABX phoneme discriminability metric \citep{schatz2016abx} as used in the Zerospeech challenge
\citep{dunbar2019zero} is based on triples of stimuli
($A, B, X$) where $A$ and $X$ belong to the same category and $B$ and $X$ belong
to different categories. The ABX error is a function of $d(A, X)$ and $d(B, X)$
where $d(\cdot, \cdot)$ is a distance metric for the representation being
evaluated:\footnote{The ABX error is thus similar to a discretized version of
the triplet loss.}
\begin{equation}
  \label{eq:abx}
  \mathrm{abx}(A, B, X) =
  \begin{cases}
    1           & \text{ if } d(A, X) > d(B, X) \\
    \frac{1}{2} & \text{ if } d(A, X) = d(B, X) \\
    0           & \text{ otherwise }
  \end{cases}
\end{equation}

The categories are determined by gold annotation: in the case of
Zerospeech they are phoneme labels. The stimuli are presented in
context in the form of minimal pairs: ($A=$/beg/$_1$, $B=$/bag/,
$X=$/beg/$_2$), where /beg/$_1$ and /beg/$_2$ are two different
utterances of this phoneme sequence. In our use case, alignments
between the gold phoneme transcriptions and the evaluated
representations are required to extract the stimuli for ABX.
Here we use the same distance metric as for RSA: Levenshtein edit distance
normalized by the length of the longer string. 


The ABX error is loosely related to the RSA score. With RSA, pairwise
distances are measured between gold representations of stimuli
(e.g.\ their phonemic transcriptions) as well as between system
representations of the same set of stimuli. The correlation
coefficient between these two sets of distance measurements is the RSA
score. With RSA there is no notion of a stimulus triple, but rather
the score reflects distances between all pairs of stimuli.  Likewise,
the representation of stimuli according to the gold standard is
typically not in the form of atomic categorical labels but can be any
representation with an associated distance (or similarity) metric.
Thus RSA can be seen as more general than ABX, while being less
controlled.


\begin{table}
  \centering
  \begin{tabular}{lcccc}
    \toprule
    Metric &  Triples  & Align   & Train & Distance  \\
    \midrule
    RSA    &     &   &    & \checkmark \\
    ABX    &  \checkmark   & \checkmark  &    & \checkmark \\
    NMI    &     & \checkmark  &    &  \\
    DC     &     & \checkmark  & \checkmark   &  \\
    \bottomrule
  \end{tabular}
  \caption{Summary of the main features of the evaluation metrics used.}
  \label{tab:metrics}
\end{table}

\paragraph{Summary of metrics}
Table~\ref{tab:metrics} summarizes the main characteristics of the
evaluation metrics described above in the context of analyzing neural
representations of speech, along the following facets: the need to
arrange input in the form of minimal-paired stimulus triples, the need
for alignment between input/codes and phonemic transcriptions, the
presence of trainable parameters, and reliance on a distance metric
between stimuli. According to these criteria, RSA and NMI are the
least restricted in their applicability, requiring only a distance
function or alignment, respectively.  In addition to a distance
metric, ABX needs minimal-paired stimulus triples to be extracted. In
addition to alignment DC has trainable parameters, and in the case of
analyzing discrete codes, it behaves like an approximation to the NMI
metric.

\subsection{Evaluation procedure}
We evaluate the induced discrete representations by applying the trained
networks on the relevant examples -- either full utterances, or speech segments
corresponding to sequences of three phonemes (triplets) -- and extract the
sequences of codes from the VQ layer. We do the same for randomly initialized
(untrained) versions of the networks, in order to provide a baseline score,
following the methodology of \citet{chrupala-etal-2020-analyzing}. In
Section~\ref{sec:results}, we include in the plots both the baseline scores and
the scores with the trained models. For the self-supervised target model we
vary only the size of the codebook in the VQ layer,\footnote{The self-supervised model has the VQ layer as part of its original
architecture, in a {\it bottleneck} composed of only one linear layer making it
hard to manipulate the level of the layer without disrupting the model.}
using sizes $2^n$ for $n \in [5, 10]$. For the visually-supervised target model
we use the same sizes and also vary the placement of the VQ layer between
one of three levels (following first, second or third GRU layer).
For both target models, each variant was trained three times with a different
random initialization; the scatter plots in Section~\ref{sec:results} show each
of these runs, as well as a LOESS fit \citep{cleveland1979robust} to each
combination of codebook size and level.

\subsection{Datasets}

\paragraph{Training target models}
Following \citet{van2020vector} we train the self-supervised model on about 15
hours of speech from over 100 speakers provided by Zerospeech
2020 \citep{dunbar_zero_2020}.\footnote{\href{https://zerospeech.com/2020/}{https://zerospeech.com/2020/}}
The visually-supervised model is trained on the Flickr8K Audio Caption Corpus
\citep{harwath2015deep,rashtchian-etal-2010-collecting}\footnote{\href{https://groups.csail.mit.edu/sls/downloads/flickraudio/}{https://groups.csail.mit.edu/sls/downloads/flickraudio/}},
which consists of 8,000 images of daily scenes each paired with five spoken captions. The
training portion of this dataset contains 6,000 images and about 34 hours of
speech.

\paragraph{Evaluation}
We encode the development captions of Flickr8K (5,000 captions) using
the encoders of the trained and untrained target models, and for each
utterance extract the sequence of codes output by the VQ layer. We
split this data in half, and use one half for training the DC, and the other 
half for computing the scores for DC, RSA and NMI.

As ABX (and one experiment with RSA) is not computed on full
utterances but on phoneme trigrams, we prepare this data by sampling
1,000 captions from the Flickr8K development set, and cutting the
audio into non-overlapping segments corresponding to a sequence of
three phonemes. We then use the ZeroSpeech code to generate
minimal-pair stimulus triples.

In order to obtain reference phonemic transcriptions we use forced alignment
with the Gentle
toolkit,\footnote{\href{http://lowerquality.com/gentle/}{http://lowerquality.com/gentle/}}
based on Kaldi \citep{Povey_ASRU2011}. This fails for a small number of
utterances, which we remove from the data.

\subsection{Repository}

The code for replicating our
experiments is available at \href{https://github.com/bhigy/discrete-repr}{https://github.com/bhigy/discrete-repr} under Apache License 2.0.

\section{Experimental results}
\label{sec:results}

Here we report experiments examining how VQ layers encode phonemes in
each target model according to different
evaluation metrics. The impact of VQ layers on performance of the
visually-supervised model in image retrieval is reported in
Section~\ref{sec:recall} of the Supplementary Material.

\subsection{Visually-supervised representations}
\label{sec:visually-supervised}

\renewcommand{\tabcolsep}{2pt}
\begin{figure*}[t!]
	\centering
	\begin{tabular}{@{} c c @{}}

		\includegraphics[trim=.2cm .2cm .6cm 0cm, clip=true, width=\columnwidth]{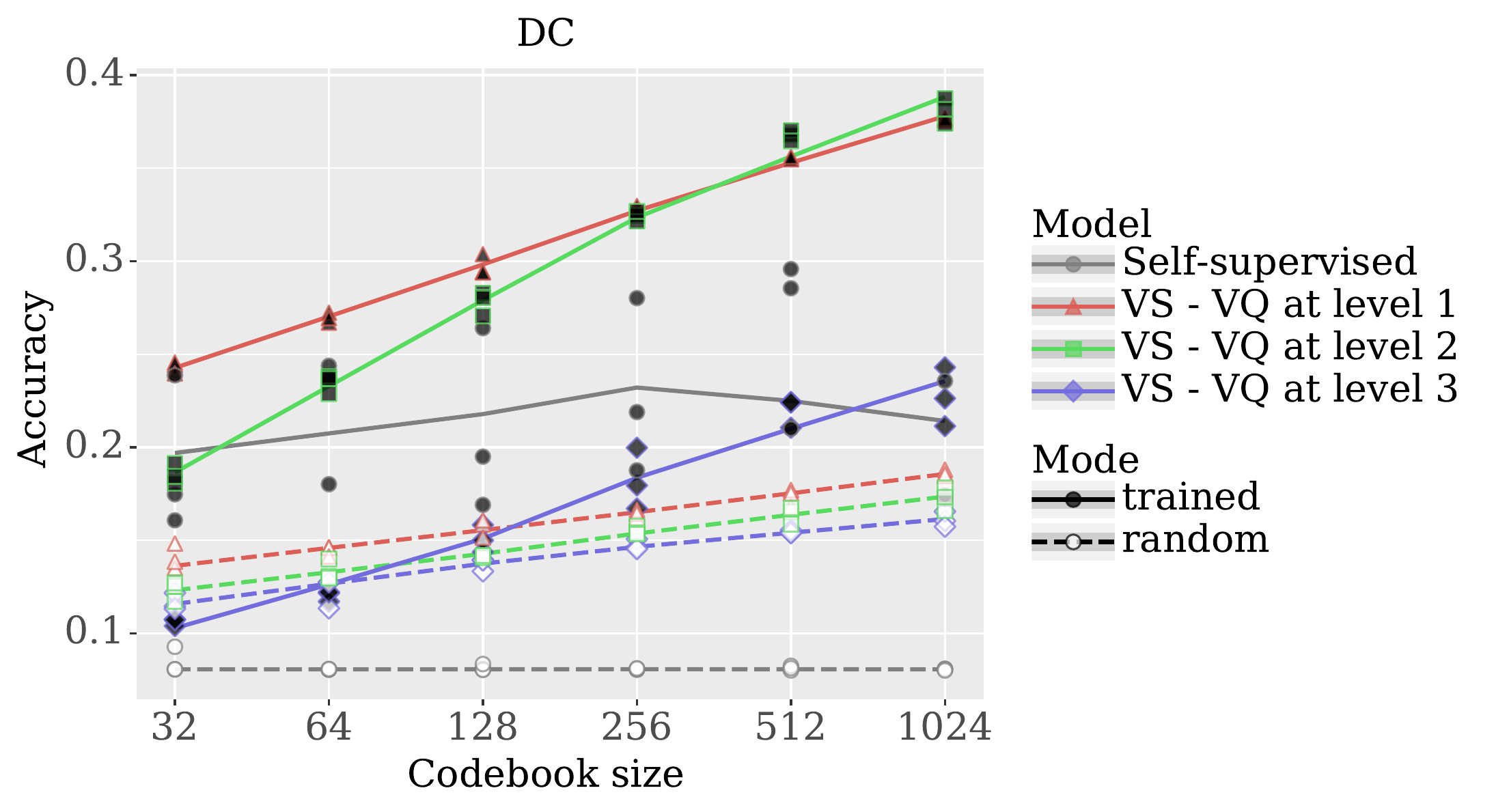} &
		\includegraphics[trim=.2cm .2cm .6cm 0cm, clip=true, width=\columnwidth]{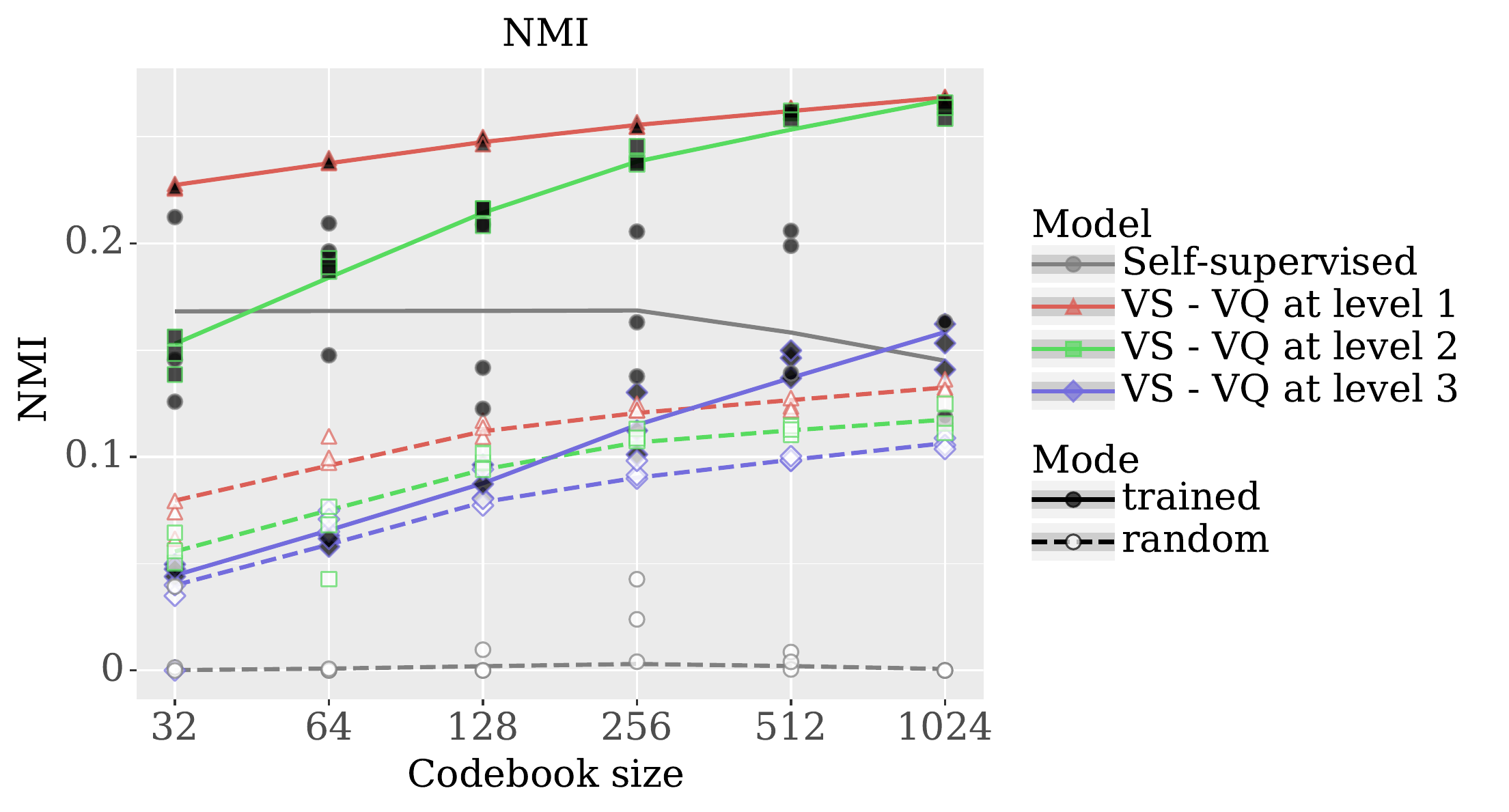} \\

		\includegraphics[trim=.2cm .2cm .6cm 0cm, clip=true, width=\columnwidth]{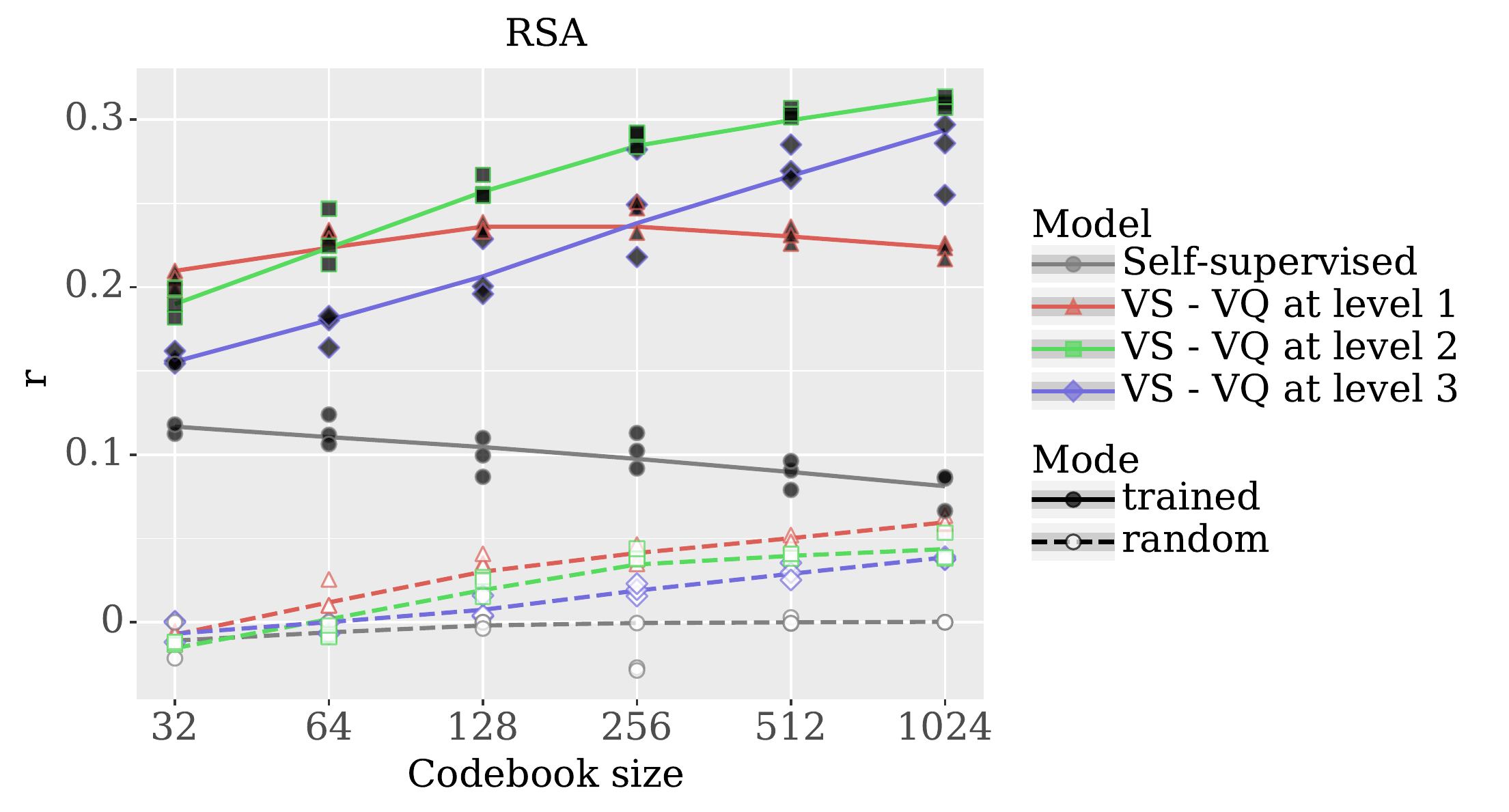} &
		\includegraphics[trim=.2cm .2cm .6cm 0cm, clip=true, width=\columnwidth]{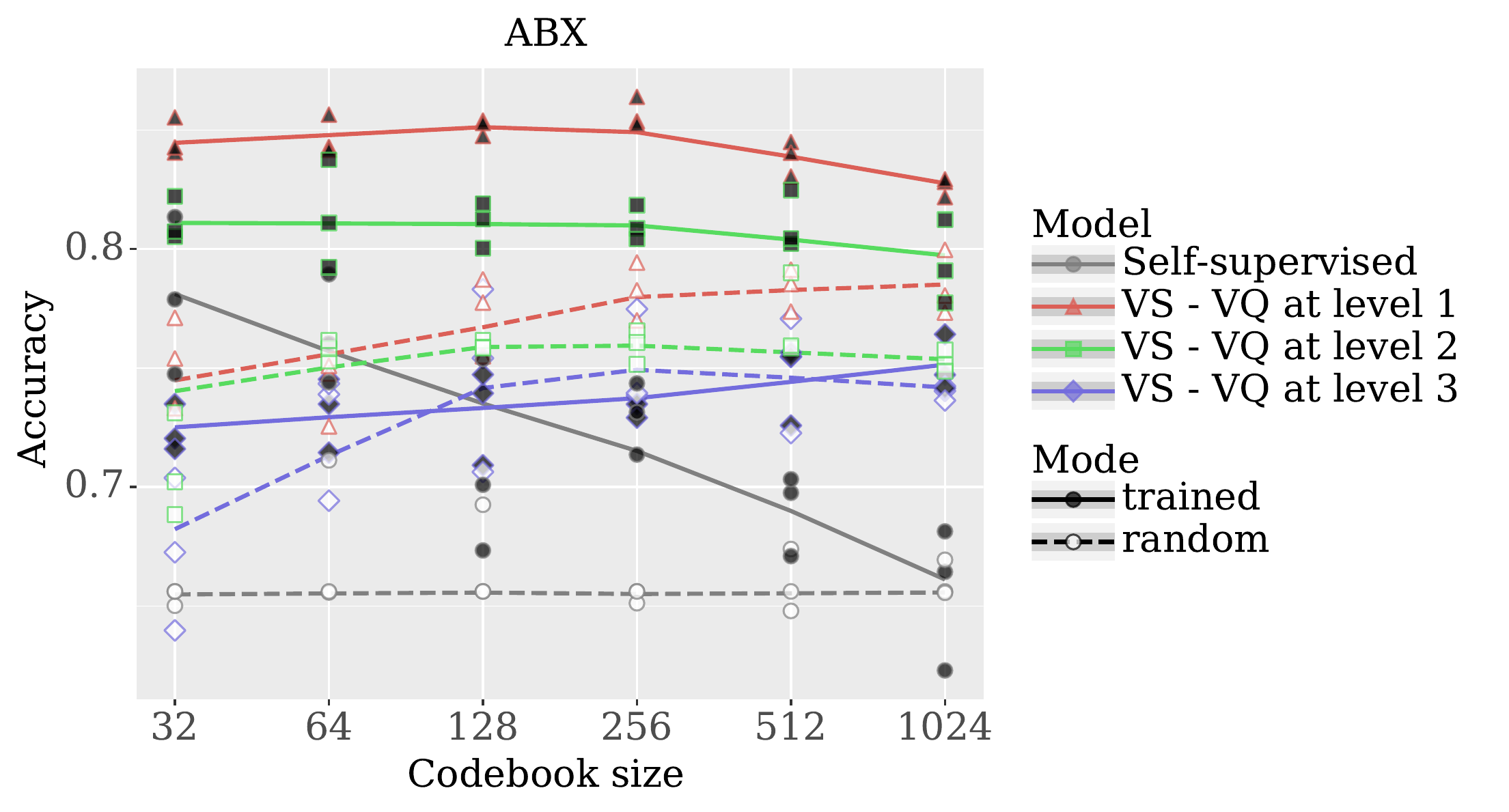} \\

	\end{tabular}
    \caption{Correspondence of codes to phonemes according to four different
        metrics as a function of codebook size, type of model, and level of VQ
        layer placement (in the visually-supervised case).\\
        {\it Top left:} Accuracy of the DC.\\
		{\it Top right:} \gls{NMI} between codes and phoneme labels.\\
		{\it Bottom left:} RSA score measured against phonemic transcriptions.\\
		{\it Bottom right:} Accuracy on the ABX task.\\
        RSA is computed on full utterances; ABX on phoneme-triple segments. DC
        and NMI are computed frame-wise on full utterances forced-aligned to the phonemic
        transcriptions. Higher scores are better for all metrics. Trained models are 
        represented by black icons and solid lines. Dashed lines and white filled icons
        represent randomly initialized models.}
	\label{fig:metrics}
\end{figure*}
\renewcommand{\tabcolsep}{6pt}

We extract codes from visually-supervised VQ models trained on
the Flickr8K data while varying VQ layer placement and codebook size. We
evaluate how much these codes correspond to  phonemes according to four
metrics:  DC, \gls{NMI}, RSA and ABX. These results are shown
in Figure~\ref{fig:metrics}.

\paragraph{DC vs.\ \gls{NMI}} As expected on theoretical grounds, diagnostic
classifiers and mutual information give very similar results. Overall, VQ
layers at level 1 and 2 perform significantly better than VQ layers at level 3
or equivalent untrained models.  Larger codebook sizes also tend to perform
better than smaller ones.

\paragraph{DC and \gls{NMI} vs.\ RSA} RSA differs from the other two
metrics on three main aspects: (i) VQ layers at level~3 perform
comparably to level~2, (ii) medium codebook sizes give better
performances for VQ layers at level 1, and (iii) untrained models show
very poor performance. The three points can be explained by the
sensitivity of RSA to the purity of the
representations. Focusing on the last point first, while representations extracted from untrained models can still contain meaningful information, it will be less explicit and
mixed with information that is not relevant for the task of
interest. A similar explanation might also hold for the first two points;
in particular, we observed that the VQ layers at level 1 retain much
more information on the speaker than the other two levels and that the encoding of speaker information increases with the size of the codebook (see
Section~\ref{sec:speaker} of the Supplementary Material for related experiments).

\paragraph{ABX vs.\ rest} ABX is the most divergent metric. While VQ
layers at level~1 and 2 still perform better, the gap with equivalent
untrained models is smaller. The effect of codebook size is also much
less pronounced and layer-specific. The difference in patterns of
results between the metrics might be due to the different testing
stimuli used by ABX versus the other three metrics: whereas DC,
\gls{NMI} and RSA are tested on full utterance audio files, ABX is
tested on small phoneme trigram files.

\paragraph{Role of stimulus size}
To disentangle the impact of the metric and the size of the stimulus
it is applied on, we run two additional experiments. First, we
re-calculate RSA using the same phoneme trigram files used for
ABX. The correlation coefficient between ABX and this version of RSA
is much stronger (see Table~\ref{tab:metric_correlation}), suggesting
that the type of stimulus used to test the model does play a role.

\begin{table}
  \centering
  \begin{tabular}{lllr}
    \toprule
    Model & RSA input & $r$ \\
    \midrule
    VS &   triplet &  0.93 \\
    VS &  complete &  0.16 \\
    SS &   triplet &  0.92 \\
    SS &  complete &  0.93 \\
  \bottomrule
  \end{tabular}

  \caption{Correlation between the ABX and RSA scores, with RSA computed
  on complete utterances and phoneme triplets, for the visually
  supervised (VS) and self-supervised (SS) models.}
  \label{tab:metric_correlation}
\end{table}

The other experiment goes in the opposite direction and brings
the ABX evaluation closer to the other three metrics.
Training the target model on full sentences but
applying it to short segments could play a role. Thus, we run an
additional set of experiments where we apply the models to full
utterances and generate the representation used in ABX by extracting
the portion of the code sequence corresponding to each phoneme trigram
from the full sequence of activations. The correlation with RSA is
still low ($0.14$) indicating that the problem is intrisic to the
evaluation relying on phoneme triplets and not train/test mismatch.
This impact of stimulus size on results is likely related to the fact
that, with very short stimuli, most normalized edit distances will be
maximum, or near maximum, and this will especially be the case for
large codebook sizes, giving very skewed and long-tailed edit distance
distributions (see details in Section~\ref{sec:ed-distro} in the
Supplementary Material).


\subsection{Self-supervised representations}

We extract codes from a number of self-supervised VQ-VAE models
trained on the Zerospeech~2019 challenge dataset with varying codebook
sizes (note that the VAE model has only one possible placement for the
VQ layer).  We compare how well these codes correspond to phonemes
according to the same four metrics, as displayed in
Figure~\ref{fig:metrics}. In contrast to what we see for
visually-supervised representations, here RSA and ABX scores are
largely consistent and suggest that a larger codebook leads to weaker
encoding of phonemic information. The effect is more pronounced for
ABX though with the largest codebook performing similarly to the
baseline. DC and \gls{NMI} are relatively
insensitive to the size of the codebook.

The self-supervised model does not show
the discrepancy observed with the visually-supervised model when
RSA and ABX scores are run on testing input of different size. This is
confirmed by the correlation coefficients between the ABX
score and RSA computed on complete utterances and phoneme triplets
shown in Table~\ref{tab:metric_correlation}.
The VQ layer in the self-supervised model only has access to a limited
context, which provides enough information to subsequently reconstruct
the input audio frame. The visually-supervised architecture on the
other hand builds a representation for the whole utterance, supported
by recurrent layers. This core difference may explain the pattern of
results we report here.

\section{Discussion}
\label{sec:discussion}

To summarize, the different metrics we compared give divergent views of the same representations.
This should not necessarily be interpreted as one metric being right while the others are flaweded.
It is likely that the different metrics account for somewhat different properties of the representation.
The differences between RSA and DC/NMI are probably related to the purity of the representation.
RSA is based on the correlation of distances between pairs of stimuli and is thus sensitive to the presence of additional information in the representation.
If the model's representation contains information that is not related
to the representation that is the target of the analysis (in our case
phonemes identification), this will be reflected in lower RSA scores.

An obvious example of such information that a model of speech is likely to encode is speaker identity.
As pointed out in Section \ref{sec:visually-supervised}, the pattern of results obtained with a classifier trained to predict speaker identity supports this view.
The encoding of speaker information could lead to comparatively lower scores with RSA, especially for level 1 and large codebooks where speaker identity is better encoded.

In general, our results suggest that different metrics might be
preferred depending on the question that one is trying to answer.
RSA scores are a better indicator of the exact match between two representations while DC/NMI better evaluate the extent to which a given information can be extracted from a model's representation, irrespective of other sources of information that might be encoded at the same time.
This could be confirmed through white-box experiments, where the different metrics would be applied to hand-crafted representations with different properties (e.g.\ in term of purity).
We leave this to future work.

The only remaining point of concern that arises from our results is the interaction between codebook size and input size.
Larger codebook sizes tend to be disadvantaged when short input segments are used as input, such as minimal pairs of phoneme triplets.
This is particularly relevant for ABX where the use of minimal pairs of phoneme triplets is a common practice.
Analysis of discrete representations using short segments should preferably be carried with diagnostic classifiers or NMI, especially if codebooks or different sizes are compared.

It is also important to highlight that the importance of this effect is dependent on the architecture and the training objective of the model, as our experiments with a self-supervised model show.

\section{Conclusion}
We compared four different metrics for discrete representations
induced by VQ layers in weakly-supervised models of spoken language,
and while the results are broadly consistent, some differences did
emerge. RSA tends to show a bigger gap between trained and untrained
models as it is more sensitive to the purity of representations with
respect to the information of interest. More surprising is the
divergent results we observe when evaluation is performed on minimal
pairs of phoneme trigrams: this is likely due to the skew of distance
distributions with large codebooks sizes.

This is an important finding as some previous work on discrete
representations focused exclusively on the ABX metric.  In contrast,
we recommend corroborating results with multiple analytical
approaches, as currently their behavior in different settings is
incompletely understood.


Overall, our findings do support the idea that vector quantization is
an effective way to induce discrete representations, and that these
correlate with symbolic representations assumed in linguistics.
However, it is worth noting that the absolute values of our metrics
measuring correspondence to phonemes are moderate at
best.\footnote{The highest RSA correlations are around $0.3$ and NMI
  values slightly above $0.25$.} It is thus important to keep in mind that these
symbolic units are not exact analogs of the concepts familiar from
linguistic theory and psycholinguistic studies.

\section*{Acknowledgements}
Bertrand Higy was supported by a NWO/E-Science Center grant
number~027.018.G03.

We would also like to thank multiple anonymous reviewers for their
useful comments which helped us improve this paper.

\bibliographystyle{acl_natbib}
\bibliography{biblio,anthology}
\clearpage
\appendix
\section{Supplementary material}

\subsection{Loss of a logistic diagnostic classifier}
\label{sec:loss-diag}
The most general measure of the amount of information about the value of a
random variable $Y$ obtained through the observation of the value of
random variable $X$ is mutual information $I(Y; X)$. Here we relate
the loss of a logistic diagnostic classifier predicting $Y$ from $X$ in
the special case where both $Y$ and $X$ are discrete, with image
$\mathcal{Y}$ and $\mathcal{X}$ respectively.

We can construct a logistic classifier which outputs the empirical
probability $\hat{P}(Y=y | X=x)$ with $y \in \mathcal{Y}$ and $x \in
\mathcal{X}$,  by using a one-hot encoding of the
categorical predictor variable $X$ as $\mathbf{x}$ and by setting the
classifier coefficients $\mathbf{W}$ as

\begin{equation}
  \label{eq:weights}
  W_{y,x} = \ln \hat{P}(Y=y | X=x).
\end{equation}
The softmax of the logistic classifier with these coefficients
simplifies to the empirical estimates of conditional probabilities of $Y$:
\begin{align}
  \label{eq:pred}
  p_{y|\mathbf{x}} & = \frac{\exp(\ln \hat{P}(Y=y|X=x))}
  {\sum_{z \in \mathcal{Y}} \exp(\ln \hat{P}(Y=z|X=x))}\\
                          & = \hat{P}(Y=y|X=x).
\end{align}
The cross entropy of the predictions of this classifer is then:
\begin{equation}
  \label{eq:cross-entropy}
  J(\mathbf{w}) =  - \frac{1}{N}\sum_{n=1}^N \ln \hat{P}(Y=y_n | X=x_n)
\end{equation}
where $y_n$ and $x_n$ are the values taken by the random variables $Y$
and $X$ for the $n^{\text{th}}$ example. The loss is equivalent to the
empirical estimate of the conditional entropy $H(Y | X)$, and related
to the mutual information between $I(Y; X)$ via:
\begin{equation}
  \label{eq:mi2}
  I(Y; X) = H(Y) - H(Y | X).
\end{equation}
To the extent that the scores of the logistic diagnostic classifier
and normalized mutual information applied to the same data are not
perfectly correlated, this would be due to the stochasticity of
training, regularization, as well as the use of accuracy rather
than cross entropy to measure performance.

\subsection{Overall performance of the visually supervised model}
\label{sec:recall}
\begin{figure}
  \centering
  \includegraphics[width=\linewidth]{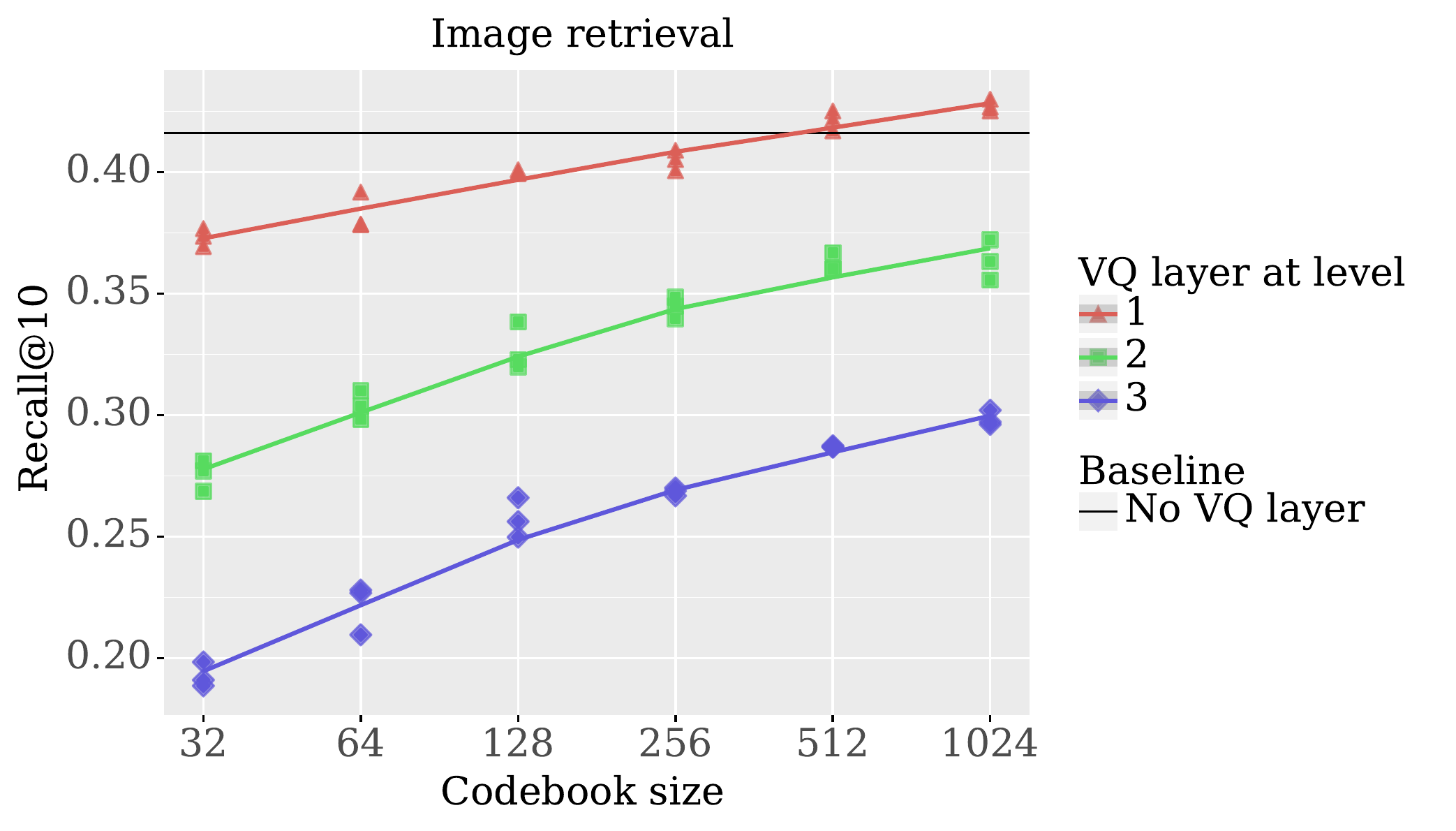}
  \caption{Recall@10 as a function of codebook size and level of
    placement.}
  \label{fig:recall_size}
\end{figure}

Since the visually-supervised model is trained and optimized for
matching images and their corresponding spoken captions, we measure the
recall@10 of retrieving the correct image for a given spoken
utterance as a function of the size of the learned codebook and the
placement of the VQ layer.
Figure~\ref{fig:recall_size} shows these results. We observed
the following patterns:
\begin{itemize}
\item Most models perform worse than a model without the VQ layer, with the
    exception of the models with the VQ layer at level 1 and a codebook of
    size $512$ or $1024$.
  \item Performance on the image retrieval task is negatively
    correlated with the level of placement of the VQ layer.
\item VQ layers with larger codebooks perform better.
\end{itemize}

\subsection{Edit distance distribution}
\label{sec:ed-distro}
Table~\ref{tab:distro} presents skew and excess
kurtosis of edit distance distributions for both target models, using
codebooks of size 32 and 1024 and trained on long and short segments,
confirming the hypothesis that the combination of short segments and
large codebooks leads to skewed distributions.

\begin{table}[t!]
  \centering
  \begin{tabular}{llrrr}
    \toprule
    Model &     Input &  Size &  Skew & Kurtosis \\
    \midrule
       SS &  complete &    32 &  0.14 &     0.21 \\
       SS &  triplet  &    32 & -0.67 &     0.55 \\
       SS &  complete &  1024 & -0.20 &     0.29 \\
       SS &  triplet  &  1024 & -2.79 &     8.99 \\
       VS &  complete &    32 &  0.04 &     0.35 \\
       VS &  triplet  &    32 & -1.66 &     3.20 \\
       VS &  complete &  1024 & -1.40 &     4.71 \\
       VS &  triplet  &  1024 & -7.92 &    87.20 \\
    \bottomrule
  \end{tabular}
  \caption{Skew and excess kurtosis of edit distance distributions, for
    the self-supervised (SS) and visually-supervised level 1 (VS) models.}
  \label{tab:distro}
\end{table}

\subsection{Speaker identification}
\label{sec:speaker}
\begin{figure}[htb]
	\centering
		\includegraphics[trim=.2cm .2cm .6cm .2cm, clip=true, width=\columnwidth]{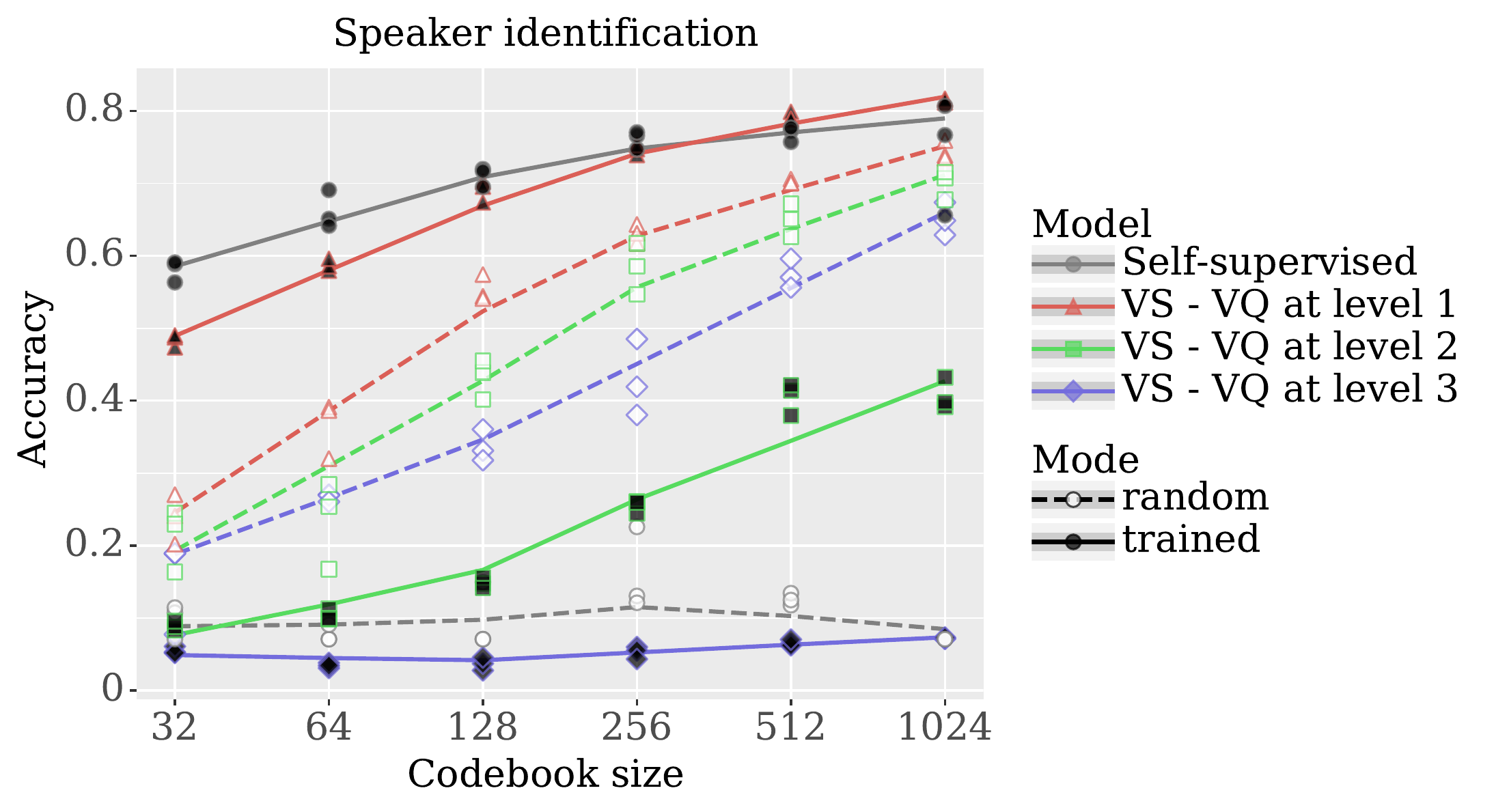}
	\caption{Accuracy of speaker classification on codebooks for the self-supervised and
          visually-supervised models. Trained models are 
          represented by black icons and solid lines. Dashed lines and white filled icons
          represent randomly initialized models.}
	\label{fig:speaker}
\end{figure}

Figure~\ref{fig:speaker} shows accuracy of diagnostic classifiers trained on
code sequences encoded as vectors of code frequencies. For visually-supervised
models, speaker identity is represented to some degree in untrained models, for
codebooks of all sizes and at all levels, and most strongly for large codebook
sizes at level~1. After training, we observe differentiated patterns for VQ
layers at level~1 versus 2 and 3. While speaker identity is emphasized in
codebooks at level~1 compared to the untrained models, it is weakened for
subsequent layers, to the point of being effectively removed at level~3. These
results indicate that VQ layers at level~1 represent speaker-dependent
information, possibly encoding acoustic rather than phonemic information. The
self-supervised models show results similar to the visually-supervised models
with the VQ layer at level 1 when trained, but capture nearly no speaker
information before training.

\end{document}